\documentclass{article}

 \usepackage[dblblindworkshop, final, nonatbib]{neurips_2025}

\usepackage[utf8]{inputenc} 
\usepackage[T1]{fontenc}    
\usepackage{hyperref}       
\usepackage{url}            
\usepackage{booktabs}       
\usepackage{amsfonts}       
\usepackage{nicefrac}       
\usepackage{microtype}      
\usepackage{xcolor}         
\usepackage{todonotes}
\usepackage[super,numbers]{natbib}

\usepackage{amsmath}
\usepackage{subcaption}

\usepackage{svg}
\usepackage{graphicx}

\usepackage{cleveref}

\usepackage{siunitx}
\usepackage{placeins}

\usepackage{float}

\title{Unveiling Latent Knowledge in Chemistry Language Models through Sparse Autoencoders}
\workshoptitle{AI for Accelerated Materials Design}

\author{
  Jaron Cohen$^{1,*}$ \quad
  Alexander G. Hasson$^{2,*}$ \quad
  Sara Tanovic$^{3,*}$ \\[0.5em]
  {\small $^{1}$Independent researcher} \\
  {\small $^{2}$Department of Oncology, University of Oxford \quad
  $^{3}$Department of Chemistry, University of Oxford} \\
  {\footnotesize \texttt{jdicohen@gmail.com} \quad
  \texttt{alexander.hasson@gtc.ox.ac.uk} \quad
  \texttt{sara.tanovic@gtc.ox.ac.uk}}
}

\begin{document}

\maketitle

\begin{abstract}

Since the advent of machine learning, interpretability has remained a persistent challenge, becoming increasingly urgent as generative models support high-stakes applications in drug and material discovery. Recent advances in large language model (LLM) architectures have yielded chemistry language models (CLMs) with impressive capabilities in molecular property prediction and molecular generation. However, how these models internally represent chemical knowledge remains poorly understood. In this work, we extend sparse autoencoder techniques to uncover and examine interpretable features within CLMs. Applying our methodology to the Foundation Models for Materials (FM4M) SMI-TED chemistry foundation model, we extract semantically meaningful latent features and analyse their activation patterns across diverse molecular datasets. Our findings reveal that these models encode a rich landscape of chemical concepts. We identify correlations between specific latent features and distinct domains of chemical knowledge, including structural motifs, physicochemical properties, and pharmacological drug classes. Our approach provides a generalisable framework for uncovering latent knowledge in chemistry-focused AI systems. This work has implications for both foundational understanding and practical deployment; with the potential to accelerate computational chemistry research.

\end{abstract}
\section{Introduction}
The intersection of artificial intelligence (AI) and chemistry has recently witnessed unprecedented advances with the emergence of foundational chemistry language models (CLMs)~\citep{moretLeveragingMolecularStructure2023, ozcelikChemicalLanguageModeling2024}. Built upon Transformer\cite{transformer} architectures, these models  have been fine-tuned for tasks in molecular property prediction and \textit{de novo} materials design, often matching or exceeding traditional approaches.\cite{chemformer, ahmad_chemberta-2_2022, edwards_translation_2022, soares_open-source_2025} Yet, these empirical successes come with a critical limitation: the models operate as ``black boxes,'' their internal decision-making processes opaque to human understanding. This interpretability challenge is particularly acute as it touches on a fundamental epistemological question: are these models learning genuine chemical principles, or are they sophisticated pattern-matching systems?

Without interpretable representations, we cannot distinguish between models that have internalised the physical laws governing molecular behaviour and those that merely memorise statistical correlations in training data. This distinction has profound implications for model generalisation, scientific discovery, and the regulatory approval of AI-assisted therapeutics. The core difficulty lies in deciphering the holistic, molecular-level vectors that represent entire chemical structures, where concepts are entangled and distributed.

\begin{figure}
    \centering
    \includegraphics[width=\linewidth]{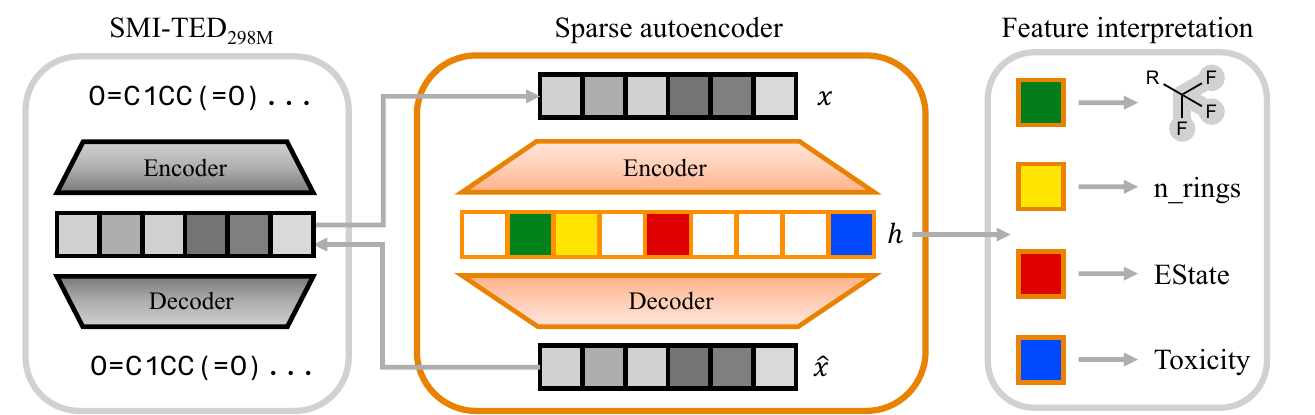}
    \caption{Overview of our workflow. Embeddings are extracted from SMI-TED and converted into features via the SAE model. These features are then interpreted to find relationships with structural and physical information.}
    \label{fig:workflow}
\end{figure}

Recent advances in sparse autoencoders (SAEs) ~\citep{olshausenSparseCodingOvercomplete1997,bricken2023monosemanticity,cunninghamSparseAutoencodersFind2023,rajamanoharanImprovingDictionaryLearning2024, templeton2024scaling} provide a promising path toward interpretability. SAEs decompose neural network activations into sparse \textit{features} that can correspond to interpretable and meaningful concepts.~\citep{bricken2023monosemanticity} However, their application to chemistry has remained unexplored. \textbf{In this work, we present the first application of SAEs to CLMs}, specifically the state-of-the-art SMI-TED foundation model\cite{soares_open-source_2025}, presenting the first systematic investigation of interpretable features within CLMs. We train SAEs on the internal representations of the model and analyse the resulting sparse features across diverse molecular datasets (\Cref{fig:workflow}). Our analysis reveals that these models develop rich, hierarchical representations of chemical knowledge, with individual features corresponding to structural motifs, physicochemical properties, and pharmacological drug classes - concepts never explicitly provided during self-supervised training. 
\paragraph{Contributions} \textbf{(1)} The first application of SAE techniques to CLMs, revealing interpretable chemical features within foundation model representations. \textbf{(2)} A novel domain-specific evaluation framework that validates chemical interpretability through molecular descriptors, substructure analysis, and functional annotations. \textbf{(3)} Demonstration of feature steering capabilities that enable causal manipulation of molecular representations while preserving chemical validity. All materials needed to reproduce our results including model weights will be made available at the time of publication.

\section{Methodology}

\subsection{Problem Formulation and Model Setup}
We formalise the problem of interpreting CLMs via sparse dictionary learning.~\citep{olshausenSparseCodingOvercomplete1997} Our central hypothesis is that a dense molecular representation vector, $\mathbf{x} \in \mathbb{R}^{d_{\text{model}}}$, can be sparsely decomposed as $\mathbf{x} \approx \sum_i h_i \mathbf{w}_i$, where $\{\mathbf{w}_i\}$ is a dictionary of interpretable feature vectors and $\mathbf{h}$ is a sparse activation vector. We extract these fixed-size vectors ($d_{\text{model}} = 768$) from the \textit{submersion layer} of the SMI-TED foundation model, which is responsible for mapping a sequence of molecular tokens into a single, fixed-size vector that represents the entire molecule.

\subsection{SAE Architecture Training and Evaluation}
To learn this decomposition, we implement a TopK SAE~\cite{Gao2024ScalingAE}. We select this architecture over traditional $L_1$-regularised approaches due to its direct control over feature sparsity and its demonstrated ability to achieve a superior fidelity-sparsity trade-off~\cite{Gao2024ScalingAE}. The encoder identifies the $k$ most active features for a given input, and the decoder then attempts to reconstruct the original vector using only this small subset.

We train our SAEs on a dataset of 5 million molecular representations extracted from SMI-TED. We curate this data from PubChem following the filtering and preparation protocol described by \citet{SMI-TED} to ensure a highly similar data distribution (see \Cref{app:pubchem_data_prep}). The training objective is to minimise reconstruction loss ($\|\mathbf{x} - \hat{\mathbf{x}}\|_2^2$) while balancing sparsity and feature utilisation.


To identify an optimal model configuration, we perform a grid search over key hyperparameters, including the dictionary expansion factor (the size of the feature dictionary relative to the input dimension, e.g., 8$\times$, 16$\times$, 32$\times$) and the sparsity level $k \in \{40, 80, 160\}$. This process allows us to map the Pareto frontier of models that optimally trade off reconstruction fidelity for sparsity. From this frontier, we select a final model that demonstrates the most promising initial signs of feature interpretability while ensuring its reconstructed vectors can still be successfully decoded back into their original, chemically valid SMILES string (see \Cref{sec:eval-methodology} for further details).

We additionally prepare the ChEMBL35~\cite{chembl}, MITOTOX~\cite{linMitoToxComprehensiveMitochondrial2021}, and MOSES~\cite{polykovskiyMolecularSetsMOSES2020} datasets as per our filtering and preparation protocol to investigate the interpretability of physicochemical, functional, and structural concepts (see \Cref{app:eval-data}).

\section{Results}
\begin{figure}[h!]
    \centering
    \includegraphics[width=1.\textwidth]{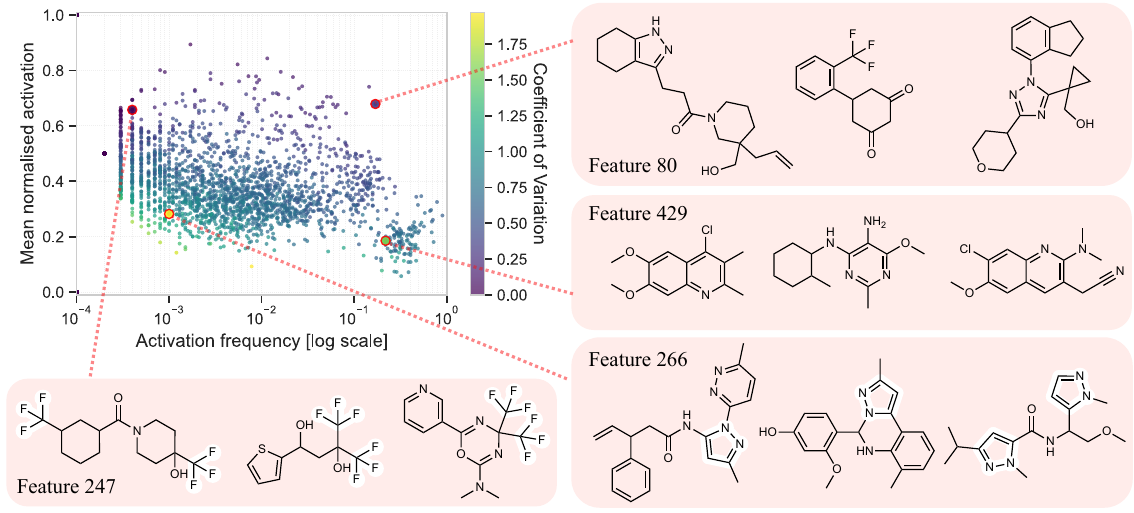}
    \caption{Feature landscape calculated on a 10k subset of the MOSES dataset. Each point represents one SAE feature, mapped according to three metrics: (1) Activation frequency (x-axis) measures how many molecules activate this feature, revealing whether it detects common or rare chemical attributes; (2) Mean normalised activation (y-axis) quantifies the typical strength when the feature activates, indicating its importance when present; (3) Coefficient of variation (colour gradient) represents consistency of activation strength, with darker points showing more consistent behaviour.
    }
    \label{fig:feature_landscape}
\end{figure}

We begin our analysis by profiling the features from our selected SAE (8$\times$ expansion, $k=80$), which is used for all subsequent investigations. We construct the \textit{feature landscape} in \Cref{fig:feature_landscape}, which maps each feature's activation frequency, intensity, and volatility. This visualisation reveals a  spectrum of feature types from specialist features (left, rare but specific) to generalist features (right, common but potentially polysemantic). For example, features 247 and 266 selectively detect specific chemical substructures (highlighted in white), activating rarely but consistently across molecules sharing these motifs, while features 80 and 429 activate frequently across structurally diverse molecules, suggesting they encode broader chemical concepts or multiple properties.

\subsection{Substructures}
\label{sec:results-sub}
We test the claim that SAEs produce an interpretable feature vocabulary by evaluating if individual features detect chemical concepts more effectively than individual neurons. A comparison of max F1 scores for 14 functional groups (\Cref{tab:functional-groups-f1}) shows that SAE features outperform neurons. This result indicates that the features form a disentangled representation, isolating specific informational components previously distributed across the latent space.

The performance gap between features and neurons is largest for motifs with low prevalence in the training data. For instance, nitrate groups appear in only 5,167 of the 5 million molecules (0.1\%). The feature for detecting this rare group achieves a perfect F1 score of 1.000, while the best neuron scores only 0.056. Large differences also exist for other low-prevalence motifs like acetylenic carbon (0.933 vs. 0.079; 0.8\% prevalence) and cyanamide (0.667 vs. 0.030; 0.1\% prevalence). These results suggest the SAE constructs new detectors from linear combinations of neurons, a necessary step when the base model avoids dedicating single neurons to rare concepts. Visualisations of top-activating molecules for these features confirm their precision, as they activate exclusively on molecules with the target substructure (Figure \ref{fig:mols_struct}).

\begin{figure}[!h]
    \centering
\includegraphics[width=\linewidth]{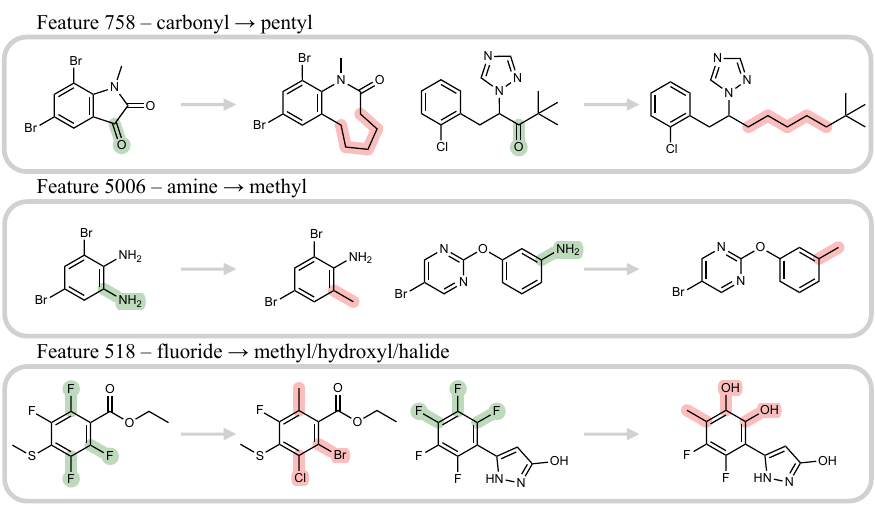}
    \caption{Examples of molecules with altered substructures highlighted before (green) and after (red) steering. Steering is performed by setting the specified feature activation to zero.}
    \label{fig:steering}
\end{figure}

To establish a causal link between a feature and its correlated motif, we perform feature steering via ablation. For a given molecule, we set the activation of a target feature to zero before decoding the molecule from its modified latent representation. Examples of this ablation experiment for three features are provided in Figure \ref{fig:steering}. This intervention produces a targeted and predictable modification; for example, ablating the feature for a carbonyl group (Feature 758) selectively removes that group from the molecular structure while preserving the core scaffold, and instead replaces it with a pentyl chain. This result provides direct evidence that the feature causally encodes the information required to generate the substructure, confirming its functional role in the model's generative process.
\vspace{-0.5em}
\vspace{-1.5em}
\subsection{Physicochemical Properties}
\label{sec:results-phys}
Beyond local substructures, SAE features also provide a more disentangled representation of global physicochemical properties. 
Comparing against raw neurons, Principal Component Analysis (PCA) components, and Non-negative Matrix Factorisation (NMF) factors, we find each descriptor correlates (Spearman’s $\rho \geq 0.3$) with 6.40 $\pm$ 3.78 features, compared to 65.77 $\pm$ 51.07 neurons, 1.08 $\pm$ 1.29 PCA components, and 3.70 $\pm$ 5.37 NMF factors. While PCA appears most parsimonious, the 100 strongest descriptor relationships map to only 3 different PCA components versus 100 unique SAE features, 5 NMF factors, and 69 unique neurons. This suggests PCA efficiently captures variance but conflates multiple chemical concepts within single dimensions. The SAE representation's lower redundancy is confirmed by the mean pairwise correlation among features (0.016 $\pm$ 0.029), which is an order of magnitude smaller than that of neurons (0.162 $\pm$ 0.122). The top three correlated features are visualised in Figure \ref{fig:mols_physico}, wherein steering of the top three activated molecules shows the causal relationship between activation and descriptor. These results demonstrate that the SAE distils the original neuron activations into a set of compact and decorrelated features, providing a more interpretable basis for aligning the model's internal representations with external physicochemical descriptors than either the original neurons or standard dimensionality reduction techniques.

\subsection{Functional Behaviour}
\label{sec:results-func}

Sections \ref{sec:results-sub} and \ref{sec:results-phys} establish that SAE features form a disentangled basis for substructures and physicochemical properties. We now investigate if these features also represent higher-level functional concepts. We first use a downstream prediction task, toxicity prediction on the MITOTOX dataset, to compare the utility and efficiency of the SAE feature basis against the original neuron basis. 

We train multiple logistic regression models on both representations and find their predictive performance to be nearly identical. Models trained on the 768 neuron activations yield a mean AUCpr of 0.603 $\pm$ 0.035, while models trained on the 6144 SAE features yield 0.606 $\pm$ 0.034, a statistically insignificant difference ($t = -0.34$, $p = 0.75$). This result confirms that the SAE decomposition is information-preserving for this task. However, the models differ in their sparsity. The logistic regression procedure identifies 213 neurons (0.277\% of variables) as significant predictors, whilst requiring only 19 SAE features (0.003\% of variables) to achieve the same performance (the top 3 of which are visualised in Figure \ref{fig:mols_toxic}). This finding suggests the SAE isolates the relevant biological signal into a more compact set of features.

We now present a case study that moves from this multi-feature abstraction to a single feature associated with a specific pharmacological mechanism. We identify a feature that activates selectively for three compounds -- CHEMBL1672485~\cite{NEMOTO20111205}, CHEMBL454618~\cite{SAKAMI20087956}, and CHEMBL368522~\cite{CLAYSON2001939} -- which span distinct chemotypes (Figure \ref{fig:mols_opioid}). While two of the molecules are very close structural analogues, the third exhibits relatively weak structural similarity to them, as measured by a Tanimoto similarity that falls within the range expected under a null distribution of random molecular pairs (see \Cref{app:func-behaviour}). There are molecules in the dataset that have higher Tanimoto similarities between all three molecules that this feature does not activate for.

All three compounds have been reported to share activity for the $\mu$-, $\kappa$-, and $\delta$-opioid receptors, according to both ZINC SEA~\cite{zinc} \textit{in silico} predictions and experimental ChEMBL bioactivity data. The maximum common substructure (MCS) across the three molecules is also present in 247 other molecules, suggesting that this feature cannot be explained by the MCS alone. This contrasts with the majority of features, which appear to correspond to small, local structural motifs.

The presence of a feature that activates across both a morphinan-like scaffold (CHEMBL1672485) and a distinct polycyclic scaffold family (CHEMBL454618/368522) suggests that the model is not solely encoding local topological similarity, but instead captures higher-order abstractions that relate to, though do not perfectly determine, shared pharmacological behaviour. The emergence of such a latent feature points toward the model's internal representations being sensitive to patterns that integrate both structural and functional information across chemotypes, rather than reflecting purely structural or purely functional groupings.

\section{Conclusion}

In this work, we demonstrate that SAEs can decompose the latent representations of a CLM into a more interpretable feature basis. This disentanglement reveals a rich landscape of features spanning local substructural motifs, global physicochemical descriptors, and high-level functional concepts. For instance, we show that the SAE isolates the signal for toxicity into a far more compact representation than the neuron basis, and that single features can group molecules by a shared pharmacological function. Crucially, we show these features are also causally relevant, enabling targeted, step-wise modifications to molecular structures through simple interventions -- a capability not afforded by the original, entangled neuron basis. This work provides evidence that the model's internal representations encode a rich hierarchy of functionally relevant chemical concepts and offers a path toward more controllable and interpretable models.

Our approach, however, has several limitations. The interpretation of features remains a presently unscaled process, and our analysis is confined to a single model architecture. The features discovered are also contingent on the training data's chemical space, and their generalisation to out-of-distribution molecules remains an open question. Furthermore, while we compare our features to the raw neuron basis, more robust baselines are needed to fully validate the decomposition's effectiveness. Future work should focus on four key directions. First, developing methods for the automated interpretation of chemical features. Second, exploring applications in AI safety, where identifying and ablating features correlated with undesirable properties (e.g., toxicity) could make models safer. Third, investigating how features behave across different model architectures and scales. This includes exploring concepts like \textit{feature splitting}~\citep{chaninAbsorptionStudyingFeature2025}, where a single feature at one SAE scale may decompose into more fundamental sub-features at a finer scale. Finally, moving beyond simple ablation to more sophisticated generative control, such as feature arithmetic, to enable multi-objective molecular optimisation~\citep{ansariDZinerRationalInverse2024}. This work provides a foundational step toward building more transparent, trustworthy, and controllable models for accelerated materials discovery.

\bibliographystyle{rsc}
\bibliography{references}

\newpage
\appendix

\makeatletter
\renewcommand \thesection{S\@arabic\c@section}
\renewcommand\thetable{S\@arabic\c@table}
\renewcommand \thefigure{S\@arabic\c@figure}
\makeatother
\setcounter{figure}{0}

\section{Background and Related Work}

\subsection{Mechanistic Interpretability}

Mechanistic interpretability is an approach to reverse engineering neural networks with the goal of understanding their internal computational mechanisms ~\citep{bereskaMechanisticInterpretabilityAI2024, raukerTransparentAISurvey2023}. A central challenge in this field is the \textit{superposition hypothesis}, which posits that models learn to represent more features than they have neurons, compressing information efficiently. This compression is thought to give rise to \textit{polysemanticity}, where a single neuron activates for multiple, seemingly unrelated concepts, thus obscuring its specific functional role. Sparse Autoencoders (SAEs) have emerged as a promising methodology for addressing this issue~\citep{olshausenSparseCodingOvercomplete1997,bricken2023monosemanticity,cunninghamSparseAutoencodersFind2023,rajamanoharanImprovingDictionaryLearning2024, templeton2024scaling}. By training an autoencoder to reconstruct a model's internal activations from a sparse, overcomplete feature dictionary, SAEs attempt to disentangle these superimposed representations. The intended outcome is the identification of \textit{putatively monosemantic} features, where each feature vector ideally corresponds to a single, human-interpretable concept, thereby rendering the model's learned knowledge more amenable to systematic analysis.

\subsection{Chemistry Foundation Models}
\label{app:chem-foundation-models}

Foundational chemistry language models (CLMs) have evolved from natural language processing and treat Simplified Molecular Input Line Entry System (SMILES) strings as the language of chemistry. State-of-the-art foundation models such as  Chemformer,\cite{chemformer} ChemBERTa,\cite{ahmad_chemberta-2_2022} MolT5,\cite{edwards_translation_2022}, and SMI-TED\cite{soares_open-source_2025} are pretrained via self-supervised learning on large databases of SMILES strings for reconstruction tasks. Models can then be fine-tuned with smaller labelled datasets towards specific chemical tasks, such as property prediction, \textit{de novo} molecular design, or retrosynthesis prediction.

\subsection{Related Work in Biological Sequence Models}
\label{app:related-work}

The application of SAEs to biological language models provides the closest precedent to our work. \citet{simon_interplm_2025} trained SAEs on ESM-2~\citep{esmfold} embeddings, successfully extracting interpretable features aligned with Swiss-Prot~\citep{uniprot} functional annotations. However, their approach fundamentally differs from ours by operating on per-residue token embeddings, enabling position-specific analysis within protein sequences. In contrast, molecular representations require handling entire chemical structures encoded as fixed-dimensional vectors, necessitating different validation strategies.
\citet{parsan_towards_2025} extended SAE analysis to protein structure prediction through ESMFold~\citep{esmfold}, demonstrating steering capabilities for structural motifs. While their steering experiments parallel our molecular steering results, the token-level granularity again distinguishes their approach. The absence of an equivalent to Swiss-Prot annotations in chemistry -- comprehensive, standardised functional labels with extensive literature evidence -- required us to develop novel validation frameworks spanning multiple chemical abstraction levels.

\subsection{SMI-TED Architecture and Training}
\label{app:smi-ted-arch-training}

SMI-TED (SMILES-based Transformer Encoder-Decoder) is a 289M parameter transformer model that novelly combines molecular token encoding with SMILES reconstruction capabilities~\citep{soares_open-source_2025}. Unlike encoder-only models, SMI-TED employs a bidirectional transformer encoder (12 layers, 768 hidden dimensions, 12 attention heads) coupled with a decoder that reconstructs complete SMILES strings from learned representations.

The model processes SMILES through molecular tokenisation, decomposing chemical structures into substructure tokens from a vocabulary of 2,993 SMILES tokens. Each token embedding $\mathbf{x}_i \in \mathbb{R}^{768}$ passes through transformer layers that incorporate rotary position embeddings (RoFormer)~\citep{su2023roformerenhancedtransformerrotary}, enabling better capture of molecular topology. Critically, SMI-TED introduces a novel submersion-immersion mechanism that maps token sequences to a unified molecular representation $\mathbf{z} \in \mathbb{R}^{768}$:
\begin{equation}
\mathbf{z} = \text{LayerNorm}\left(\text{GELU}(\mathbf{XW}_1 + \mathbf{b}_1)\right)\mathbf{W}_2,
\end{equation}
where $\mathbf{X} \in \mathbb{R}^{L \times 768}$ represents the sequence of $L$ token embeddings, $\mathbf{W}_1 \in \mathbb{R}^{L \times 768}$, $\mathbf{b}_1 \in \mathbb{R}^{768}$, and $\mathbf{W}_2 \in \mathbb{R}^{768 \times 768}$. This latent representation enables both molecular property prediction and full SMILES reconstruction - a capability that enforces learning of complete chemical information.

\subsection{Sparse Autoencoders Formulation}
\label{app:sae-formulation}

An SAE can be generally defined by its encoder and decoder functions:
\begin{align}
    \left.\begin{aligned}
    \text{Encoder: } &\quad \mathbf{h} = f_{\text{enc}}(\mathbf{x}) \in \mathbb{R}^n \\
    \text{Decoder: } &\quad \hat{\mathbf{x}} = f_{\text{dec}}(\mathbf{h}) \in \mathbb{R}^d
    \end{aligned}\right\} \text{SAE}(\mathbf{x}) = f_{\text{dec}}(f_{\text{enc}}(\mathbf{x})) = \hat{\mathbf{x}}
\end{align}
where $f_{\text{enc}}: \mathbb{R}^d \to \mathbb{R}^n$ maps the input $\mathbf{x}$ to a high-dimensional latent space ($n \gg d$) and $f_{\text{dec}}: \mathbb{R}^n \to \mathbb{R}^d$ reconstructs it. The model is trained by minimising a general loss function that balances reconstruction fidelity with a sparsity-inducing term~\citep{bussmannBatchTopKSparseAutoencoders2024a}:
\begin{align}
    \mathcal{L}(\mathbf{x}) = \underbrace{\|\mathbf{x} - \hat{\mathbf{x}}\|_2^2}_{\text{Reconstruction}} + \underbrace{\lambda \mathcal{S}(\mathbf{h})}_{\text{Sparsity}}+ \underbrace{\alpha \mathcal{L}_{\text{aux}}}_{\text{Auxiliary}}
\end{align}

This general formulation captures most SAE architectures through their specific definitions of the encoder $f_{\text{enc}}$, sparsity penalty $\mathcal{S}(\mathbf{h})$, and inclusion of an auxiliary loss.

In almost all cases, the decoder is a linear transformation $f_{\text{dec}}(\mathbf{h}) = \mathbf{W}_{\text{dec}} \mathbf{h} + \mathbf{b}_{\text{pre}}$, where $\mathbf{W}_{\text{dec}} \in \mathbb{R}^{d \times n}$. The architectures differ primarily in their encoder and loss configuration.

\paragraph{Traditional L1 SAEs~\citep{cunninghamSparseAutoencodersFind2023}:} the encoder is $f_{\text{enc}}(\mathbf{x}) = \text{ReLU}(\mathbf{W}_{\text{enc}}(\mathbf{x} - \mathbf{b}_{\text{pre}}) + \mathbf{b}_{\text{enc}})$. Sparsity is encouraged via the L1-norm ($\mathcal{S}(\mathbf{h}) = \|\mathbf{h}\|_1$), and an auxiliary loss is generally not used ($\alpha=0$).

\paragraph{TopK SAEs~\citep{Gao2024ScalingAE}:} the encoder is $f_{\text{enc}}(\mathbf{x}) = \text{TopK}(\text{ReLU}(\mathbf{W}_{\text{enc}}(\mathbf{x} - \mathbf{b}_{\text{pre}}) + \mathbf{b}_{\text{enc}}), k)$, where $\text{TopK}$ sets all but the $k$ largest elements to zero. The explicit sparsity penalty is absent ($\lambda=0$), and an auxiliary loss is often included ($\alpha > 0$) to encourage feature utilisation and prevent dead features.

\paragraph{Matryoshka SAEs~\citep{bussmann2025learningmultilevelfeaturesmatryoshka}:} the key innovation is in the reconstruction loss. Instead of a single term, the loss is a sum over multiple nested dictionaries of increasing size, $\mathcal{M} = \{m_1, m_2, \dots, m_n\}$. For each size $m \in \mathcal{M}$, a partial reconstruction $\hat{\mathbf{x}}^{(m)}$ is computed using only the first $m$ latents and the corresponding columns of the decoder matrix:
    \begin{equation}
        \hat{\mathbf{x}}^{(m)} = \mathbf{W}_{\text{dec}, 0:m} \mathbf{h}_{0:m} + \mathbf{b}_{\text{pre}}
    \end{equation}
The total reconstruction loss is the sum of the errors for each of these partial reconstructions:
    \begin{equation}
        \mathcal{L}_{\text{recon}} = \sum_{m \in \mathcal{M}} \|\mathbf{x} - \hat{\mathbf{x}}^{(m)}\|_2^2
    \end{equation}
This objective forces earlier features to learn general concepts, while later features can specialise.
\section{Experimental Setup}
\label{app:experimental-setup}
\subsection{Data Curation}
\label{app:data-curation}
\subsubsection{PubChem Training Data}
\label{app:pubchem_data_prep}
Following the exact curation procedure described by \citet{SMI-TED}, we independently filtered over 122 million molecules from PubChem (July 2025) using their reported filtering process: molecular validity verification, canonicalisation, deduplication; additionally we performed desalting before deduplication. After filtering, the dataset contained approximately 91 million molecules, from which we uniformly sample 5 million molecules for SAE training. While we cannot guarantee identical overlap with SMI-TED's training data (as it was not publicly released), following the same curation procedure should yield a dataset with highly similar distributional properties. 

\subsubsection{Evaluation Data}
\label{app:eval-data}
The MOSES, ChEMBL35 and MITOTOX datasets were prepared identically to the PubChem training data (see: \Cref{app:pubchem_data_prep}). The final dataset sizes after preprocessing are provided in each section below.

\paragraph{MOSES} MOSES~\citep{polykovskiyMolecularSetsMOSES2020} is a diverse representation of drug-like small molecule space that includes molecules optimised for drug development. We combine the MOSES test sets (N = 352,299) into a single evaluation dataset for assessing feature generalisation beyond the training distribution, providing a test of whether features learned on PubChem generalise to pharmaceutically relevant chemistry. 

\paragraph{ChEMBL} ChEMBL is a data source of literature validated functional and physicochemical annotations of molecules. In order to investigate functional relationships, we retrieved all small molecules ($<$ 500 Da) from ChEMBL35 ~\citep{chembl}. Using the same preprocessing steps as the PubChem dataset, the resultant dataset contained 1,981,621 molecules. Calculated properties such as LogD, and functional measures such as binding affinity and targets were retrieved.

\paragraph{MITOTOX}
 MITOTOX~\citep{linMitoToxComprehensiveMitochondrial2021} is a dataset of small molecules with related mitochondrial toxicity annotations. A prepared subset was retrieved as per chemeleon-tox~\citep{githubGitHubJacksonBurnschemeleon_tox}, and resulted in 3,742 molecules; 529 of which were labelled toxic (14.1\%).

\paragraph{Physicochemical Descriptors and Substructures}
For all molecules Mordred 2D descriptors~\citep{githubGitHubJacksonBurnsmordredcommunity,moriwakiMordredMolecularDescriptor2018} (N = 1613) were calculated. For all molecules Atom Invariant Morgan fingerprints (radius = 2, use\_chirality = True) were used to generate their substructure sets. For calculating the Tanimoto similarity distribution, a 4096-bit fingerprint size was used.

\subsection{SAE Training Details}
\label{app:sae-training-config}
We build atop the implementation of TopK SAEs by \citet{marks2024dictionary_learning}, originally developed for large language model interpretability. The official SMI-TED model is available on the Hugging Face Hub \cite{huggingface_materials_smi_ted}. Each sweep configuration required approximately 3 GPU hours on an NVIDIA L4 GPU. During training, all molecular representations are normalised to have a unit mean squared norm as per ~\citep{lieberum_gemma_2024} .

\begin{table}[h!]
\centering
\caption{Hyperparameter configuration for TopK SAE Training. Values in parentheses represent the grid search range.}
\label{tab:hyperparameters}
\begin{tabular}{ll}
\toprule
\textbf{Hyperparameter} & \textbf{Value(s)} \\
\midrule
Dictionary Size Multiplier & (8, 16, 32) \\
Learning Rate (lr) & 0.0001 \\
Top-K (k) & (40, 80, 160) \\
AuxK Alpha ($\alpha$) & 0.03125 \\
Training Epochs & 80 \\
Batch Size & 256 \\
Warmup Steps Fraction & 0.05 \\
\bottomrule
\end{tabular}
\end{table}

The SAE training uses the Adam optimiser~\citep{kingmaAdamMethodStochastic2014} with $\beta_1=0.9$, $\beta_2=0.999$, and a learning rate schedule that includes linear warmup followed by linear decay beginning at 80\% of total training steps.

The Warmup Steps Fraction parameter controls the proportion of total training steps during which the learning rate gradually increases from zero to its target value, implementing a learning rate warmup schedule that helps stabilise early SAE training and improve convergence.

The AuxK Alpha parameter controls the weighting coefficient for the
auxiliary loss term in TopK SAE training, which encourages the model to use a broader set of features beyond just the top-k activations to improve feature diversity and reduce dead neurons.

\subsection{Evaluation Methodology}
\label{sec:eval-methodology}

A core challenge in mechanistic interpretability is that standard SAE training metrics are only proxies for the true goal of discovering human-interpretable features, and developing robust interpretability metrics remains an open problem~\citep{bereskaMechanisticInterpretabilityAI2024}.

Our evaluation strategy is therefore twofold. First, we assess the SAE's reconstruction fidelity - its ability to preserve the essential chemical information from the original model. Second, we evaluate the chemical meaningfulness of the learned features using a hierarchical framework designed to probe for specific, domain-relevant concepts.

\subsubsection{Reconstruction Fidelity Metrics}

To measure how well the SAE's reconstructed vector, $\hat{x}$, preserves the information in the original vector, $x$, we use a combination of standard and domain-specific metrics.

Our primary measure is \textbf{functional fidelity}, defined as the success rate of decoding an SAE-reconstructed vector back into a chemically valid and equivalent SMILES string. This is a particularly stringent criterion, as minor errors can render a SMILES string invalid. We measure this at two levels: \textbf{strict accuracy} (exact canonical string matching) and \textbf{stereo accuracy} (chemical equivalence ignoring stereochemistry). High functional fidelity thus provides direct evidence that our SAE preserves the essential chemical information required by the foundation model. We supplement this with several standard metrics, which we define in \Cref{tab:recon-metrics}.

\begin{table}[h!]
\centering
\caption{Standard metrics used to evaluate SAE reconstruction fidelity.}
\label{tab:recon-metrics}
\begin{tabular}{@{}p{8cm}c@{}}
\toprule
\textbf{Metric \& Description} & \textbf{Formula} \\
\midrule
\textbf{L2 Reconstruction Loss} & \\
The primary training objective. & $\|\mathbf{x}_i - \hat{\mathbf{x}}_i\|_2^2$ \\
\addlinespace

\textbf{Fraction of Variance Explained} & \\
Quantifies the variance of the original vector captured by the reconstruction. & $1 - \frac{\text{Var}(\mathbf{x}_i - \hat{\mathbf{x}}_i)}{\text{Var}(\mathbf{x}_i)}$ \\
\addlinespace
\textbf{Fraction Alive} & \\
The percentage of SAE features that activate on at least one molecule in the validation set. & - \\
\addlinespace
\textbf{Delta Loss} & \\
Measures the preservation of the original model's loss landscape. $\mathcal{L}_{\text{SMI-TED}}$ is the original model's loss, composed of a token prediction cross-entropy term and an embedding MSE term~\citep{soares_open-source_2025}. A low $\Delta\mathcal{L}$ indicates high preservation. & $\Delta\mathcal{L} = \mathcal{L}_{\text{SMI-TED}}(\hat{\mathbf{x}}_i) - \mathcal{L}_{\text{SMI-TED}}(\mathbf{x}_i)$ \\
\bottomrule
\end{tabular}
\end{table}

\subsubsection{Framework for Evaluating Chemical Meaning}
\label{app:chem-meaning}

To systematically probe for chemical meaning, we validate features against hierarchical framework designed to capture the multi-scale nature of molecular properties. This framework consists of three categories: \textbf{substructural patterns}, which are local, discrete motifs such as functional groups and ring systems; (2) \textbf{physicochemical properties}, which are global, often continuous, properties emerging from the entire structure, like molecular weight or topological polar surface area, or systematic, high-dimensional features that encode topological, electronic, and geometrical information such as Mordred descriptors; and (3) \textbf{functional relationships}, which are abstract classifications, such as pharmacological drug class, that may not be apparent from simple structural similarity alone.

We use a logistic regression framework with a fixed random seed and class-balanced weights for two separate analyses. All models are trained and evaluated using a 5-fold cross-validation scheme.

\paragraph{Substructure Detection} To evaluate how well individual features and neurons detect specific functional groups, we train a separate logistic regression model for each feature and each neuron. The model's task is to predict the presence or absence of a single functional group. We report the maximum F1 score achieved across the validation folds as the primary performance metric.
\begin{align}
    \left.
    \begin{aligned}
        \text{Precision} &= \frac{\text{True Positives}}{\text{True Positives}+\text{False Positives}} \\[2ex]
        \text{Recall} &= \frac{\text{True Positives}}{\text{True Positives} + \text{False Negatives}}
    \end{aligned}
    \right\}
    \quad \text{F1} &= 2 \cdot \frac{\text{Precision} \cdot \text{Recall}}{\text{Precision} + \text{Recall}}
\end{align}

\paragraph{Physicochemical}
We calculate the pairwise Spearman correlation coefficient between each feature- and neuron-descriptor pair, and apply identical analysis to PCA components and NMF factors (the latter requiring a shift transformation to ensure non-negativity) extracted using scikit-learn~\citep{scikit-learn}. Where the set of descriptors is the Mordred 2D descriptors. Correlations with corresponding $p$-values $< 0.05$ were considered significant. The absolute value of the Spearman's $\rho$ was used to rank the strength of the relationship.

\paragraph{Toxicity Prediction} To assess the representational efficiency for a downstream task, we train two multiple logistic regression models to predict toxicity on the MITOTOX dataset. One model uses the complete set of SAE features as input, while the other uses all raw neuron activations. We identify inputs that are significantly predictive of toxicity by selecting model coefficients with a $p$-value $<0.05$. The performance of each model is given by the area under the precision-recall curve ($\text{AUC}_{\text{pr}}$).

\section{Feature Characterisation}
\label{app:feature-characterisation}

\subsection{Sparsity-Fidelity Trade-offs}

To systematically evaluate SAE configurations, we analysed reconstruction quality and feature utilisation across a representative subset of 10,000 molecules uniformly sampled from the MOSES dataset (described in \Cref{app:eval-data}).

\begin{figure}[h!]
    \centering
    \begin{subfigure}[t]{0.49\textwidth}
        \centering
        \includegraphics[width=\textwidth]{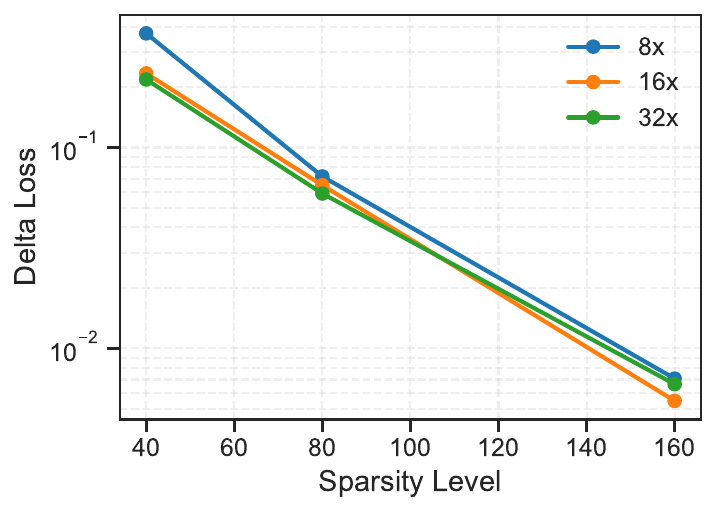}
        \caption{Delta Loss vs. Sparsity Level. This plot shows the change in the foundation model's loss when using the SAE-reconstructed vector instead of the original. Lower values indicate better preservation of the original model's loss landscape. As expected, reconstruction fidelity improves (delta loss decreases) as the number of active features increases.}
        \label{fig:delta_loss_sparsity}
    \end{subfigure}
    \hfill 
    \begin{subfigure}[t]{0.49\textwidth}
        \centering
        \includegraphics[width=\textwidth]{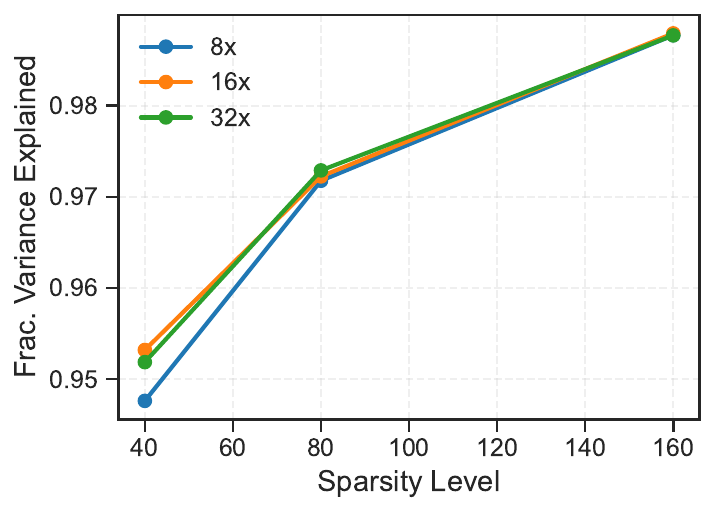}
        \caption{Fraction of Variance Explained vs. Sparsity Level. This plot quantifies the proportion of the original activation vector's variance captured by the SAE's reconstruction at different sparsity levels. Higher values indicate a more faithful reconstruction of the original vector. As expected, reconstruction fidelity improves as the number of active features increases.}
        \label{fig:frac_variance_sparsity}
    \end{subfigure}
    \caption{Sparsity-Fidelity Trade-off Across SAE Configurations.}
    \label{fig:sparsity_fidelity_standard_metrics}
\end{figure}

\begin{figure}[h!]
    \centering
    \includegraphics[width=0.9\textwidth]{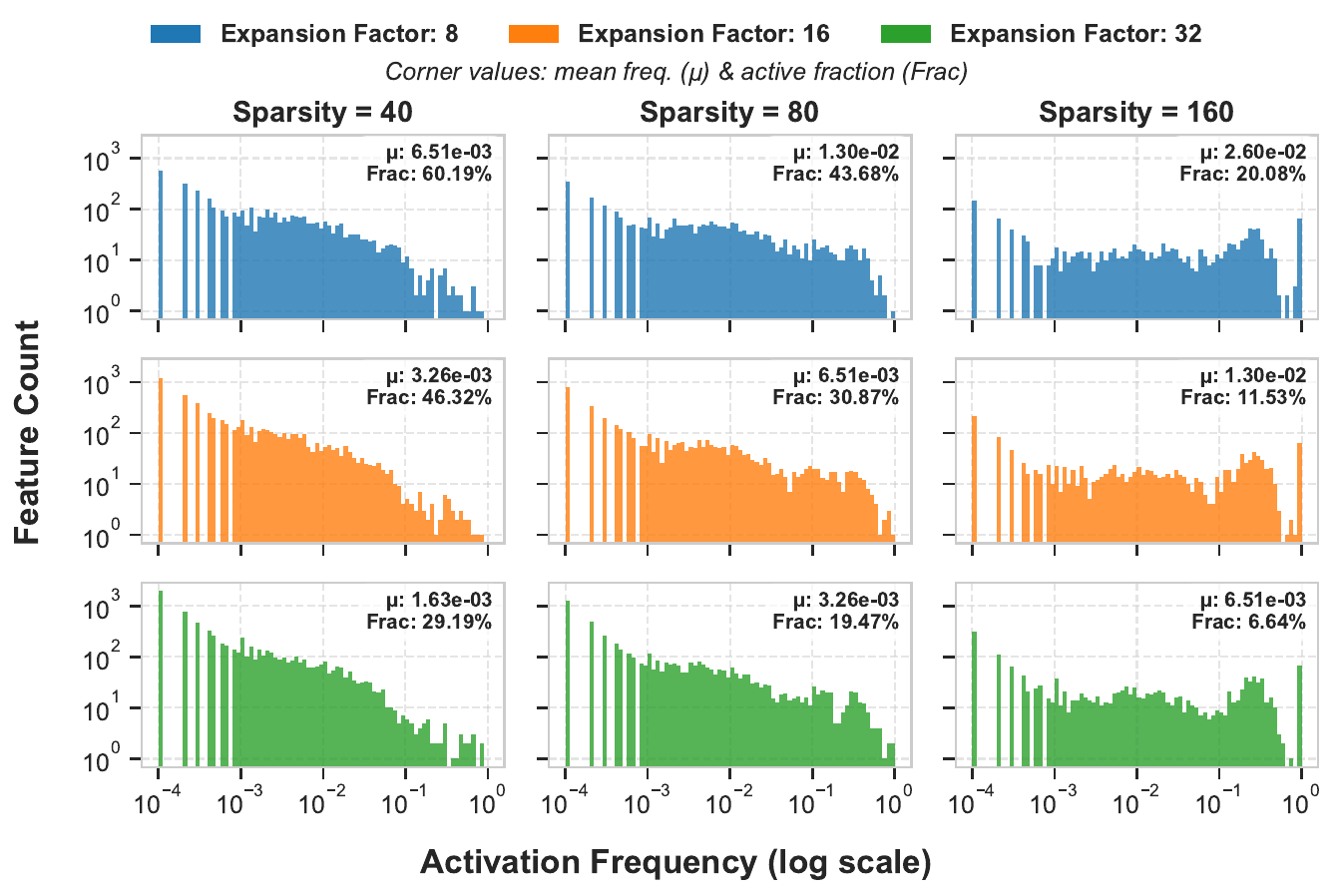}
    \caption{Feature activation frequency distributions across SAE hyperparameter configurations. Each subplot shows the histogram of activation frequencies for individual SAE features, organised by expansion factor (rows) and sparsity level (columns). Histograms use logarithmic binning and scaling to visualise the characteristic heavy-tailed distribution of feature activations.}
    \label{fig:feature_density_distribution}
\end{figure}

\begin{figure}[h!]
    \centering
    \begin{subfigure}[t]{0.48\textwidth}
        \centering
        \includegraphics[width=\textwidth]{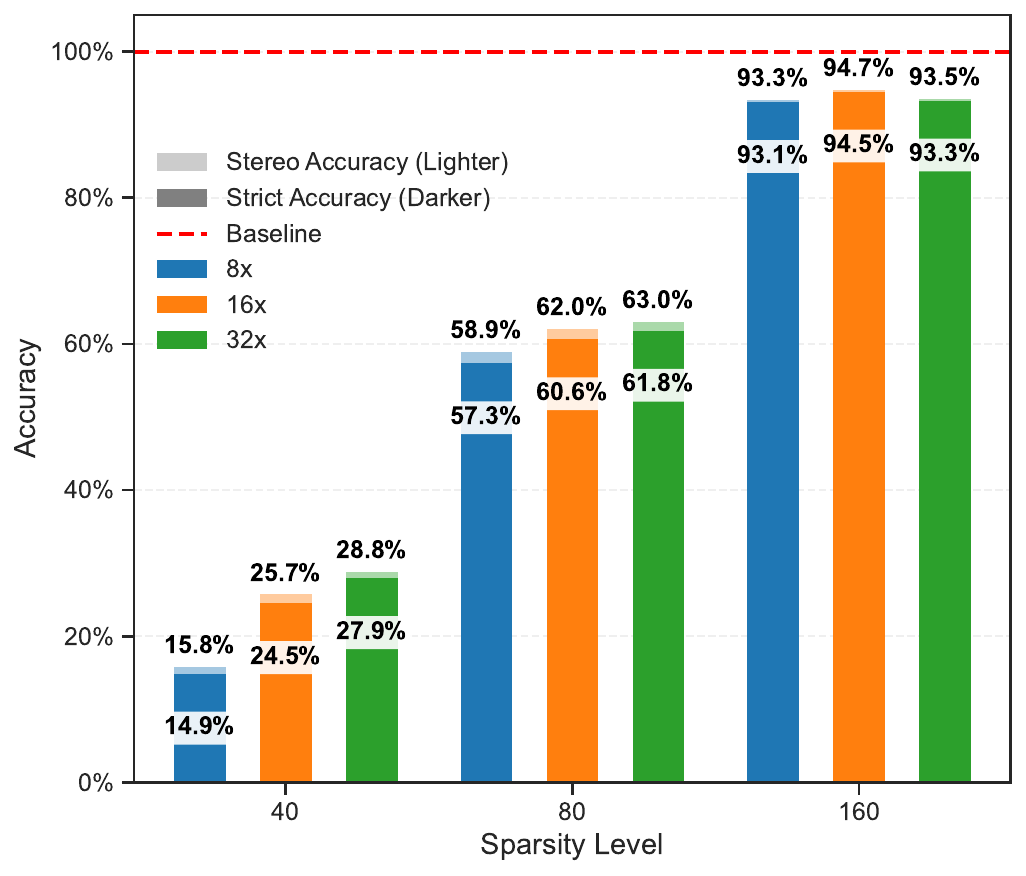}
    \caption{Comparison of SMILES reconstruction accuracy across different SAE hyperparameter configurations. Dark bars show strict accuracy (exact string matching), while light bars show stereo accuracy (chemical equivalence ignoring stereochemistry). The red dashed line indicates baseline model performance without SAE reconstruction. Results demonstrate how reconstruction accuracy varies with SAE architecture parameters, with higher expansion factors generally maintaining better accuracy recovery across sparsity levels.}
    \label{fig:accuracy_recovered}
    \end{subfigure}
    \hfill 
    \begin{subfigure}[t]{0.48\textwidth}
        \centering
        \includegraphics[width=\textwidth]{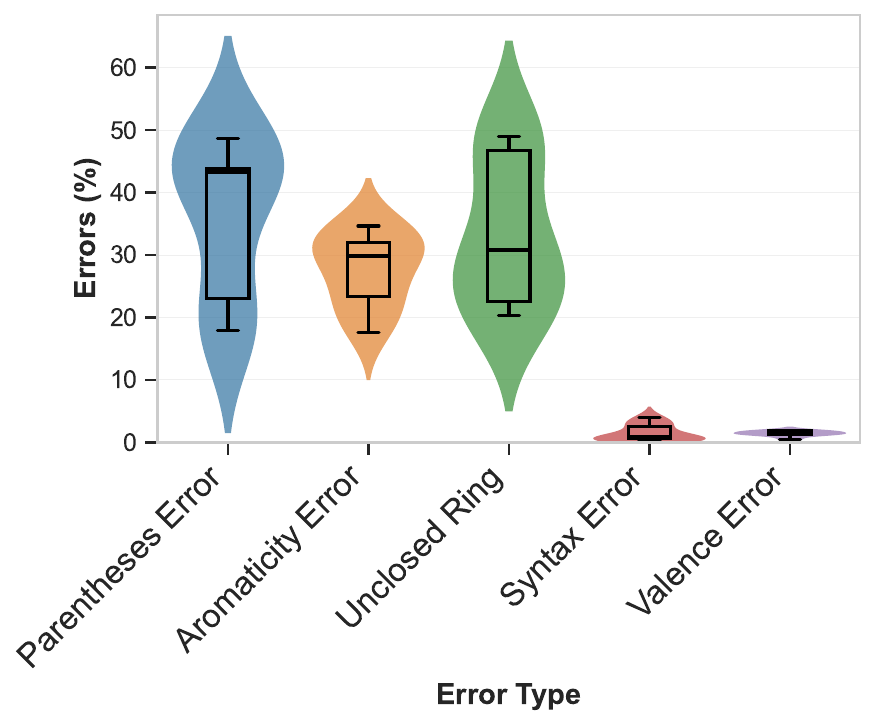}
    \caption{
    Distribution of SMILES reconstruction error types across SAE hyperparameter configurations. Box plots show the percentage distribution of different error categories (valence errors, aromaticity errors, bond duplication, unclosed rings, parentheses errors, and syntax errors) that occur when SAE-reconstructed embeddings are decoded back to SMILES strings. Error categories are classified using standardised RDKit parsing error analysis to understand how SAE reconstruction affects different aspects of molecular structure representation.}
    \label{fig:error_distribution}
    \end{subfigure}
    \caption{Impact of SAE Hyperparameters on SMILES Reconstruction Fidelity.}
    \label{fig:accuracy_recovered_error_distribution}
\end{figure}

Figures~\ref{fig:sparsity_fidelity_standard_metrics}--\ref{fig:accuracy_recovered_error_distribution} characterise the trade-off space across expansion factors ($8\times$, $16\times$, $32\times$) and sparsity levels (40, 80, 160), revealing distinct operating regimes for each configuration.

The delta loss curves (\Cref{fig:delta_loss_sparsity}) demonstrate an exponential relationship between sparsity and downstream task preservation. Interestingly, expansion factor shows minimal impact on delta loss at matched sparsity levels, suggesting that dictionary size primarily affects feature granularity rather than reconstruction quality.

The plot of fraction of variance explained (\Cref{fig:frac_variance_sparsity}) shows that reconstruction fidelity increases with $k$, but with diminishing returns. This analysis informed our selection of $k=80$ for the primary model, as it captures a high proportion of the original vector's variance ($\approx0.972$) while a further doubling of $k$ to 160 yields only a marginal improvement (to $\approx0.987$).

\FloatBarrier

\subsection{SMILES Reconstruction Analysis}

\FloatBarrier
\newpage
\section{Supplementary Results}
\subsection{Substructures}

A total of 14 common functional groups~\cite{daylightDaylightgtSMARTSExamples} were retrieved in SMARTS format to provide a range of substructures present in the dataset at varying prevalence. The SMARTS strings were used to identify the presence/absence of the functional group. Some molecules contain the same functional group multiple times, or multiple functional groups.

Steering was performed by intervening on the specified feature values, and setting the feature activation to 0.

\begin{table}[h!]
\centering
\caption{Maximum F1 scores and prevalence in 5M PubChem for various functional groups}
\label{tab:functional-groups-f1}

\begin{tabular}{lccS[table-format=2.3]}
\toprule
\textbf{Functional Group} & \multicolumn{2}{c}{\textbf{Maximum F1}} & {\textbf{Prevalence (\%)}} \\
\cmidrule(lr){2-3}
 & \textbf{Features} & \textbf{Neurons} & \\
\midrule
Alkyl Carbon & \textbf{0.945} & 0.938 & 88.286 \\
Acetylenic Carbon & \textbf{0.933} & 0.079 & 0.758 \\
Carbonyl group, High specificity & \textbf{0.745} & 0.735 & 51.197 \\
Cyanamide & \textbf{0.667} & 0.030 & 0.086 \\
Ether & \textbf{0.792} & 0.655 & 36.245 \\
Primary amine, not amide & \textbf{0.697} & 0.431 & 8.613 \\
Azo nitrogen & \textbf{0.706} & 0.071 & 0.637 \\
Nitrate & \textbf{1.000} & 0.056 & 0.103 \\
Hydroxyl & \textbf{0.838} & 0.654 & 39.569 \\
Peroxide groups & \textbf{0.667} & 0.043 & 0.224 \\
Phosphoric acid group 1 & \textbf{0.571} & 0.075 & 0.448 \\
Thiol & \textbf{0.900} & 0.135 & 0.982 \\
Sulfide & \textbf{0.700} & 0.344 & 7.907 \\
Chloride (Carbon-attached) & \textbf{0.802} & 0.533 & 21.258 \\
\bottomrule
\end{tabular}
\end{table}

\begin{figure}[h!]
    \centering
    \includegraphics[width=\linewidth]{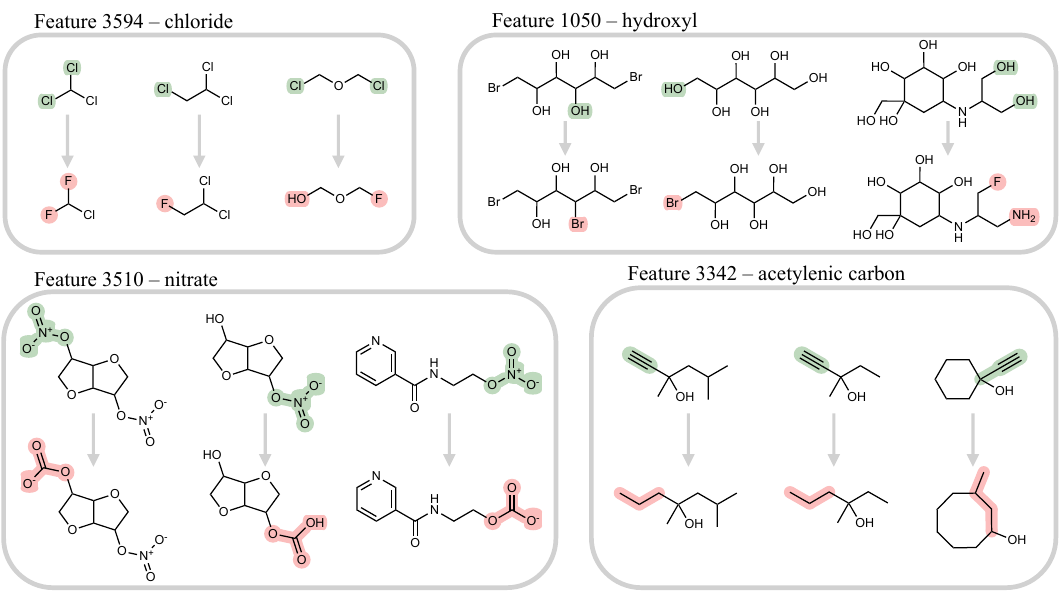}
    \caption{Top three activated molecules for the top correlated feature for functional groups chloride, hydroxyl, nitrate, and acetylenic carbon. Molecules are then steered by setting the specified feature activation to 0, and the altered substructure is highlighted before (green) and after (red) steering.}
    \label{fig:mols_struct}
\end{figure}

\FloatBarrier
\newpage
\subsection{Physicochemical Properties}
The top 3 feature-descriptor relationships, ranked by Spearman correlation, were selected. The top 3 molecules which had the highest activations for each corresponding feature were retrieved. The feature activation was set to zero, and the corresponding descriptor was recalculated. These were (with Spearman's $\rho$): StsC; sum of tsC ($\rho$ = 0.89), SMR\_VSA7; MOE MR VSA Descriptor 7 (3.05 $\leq$ x $<$ 3.63) ($\rho$ = 0.85)and Xch-3d; 3-ordered Chi chain weighted by sigma electrons ($\rho$ = 0.75).

\begin{figure}[h!]
    \centering
    \includegraphics[width=\linewidth]{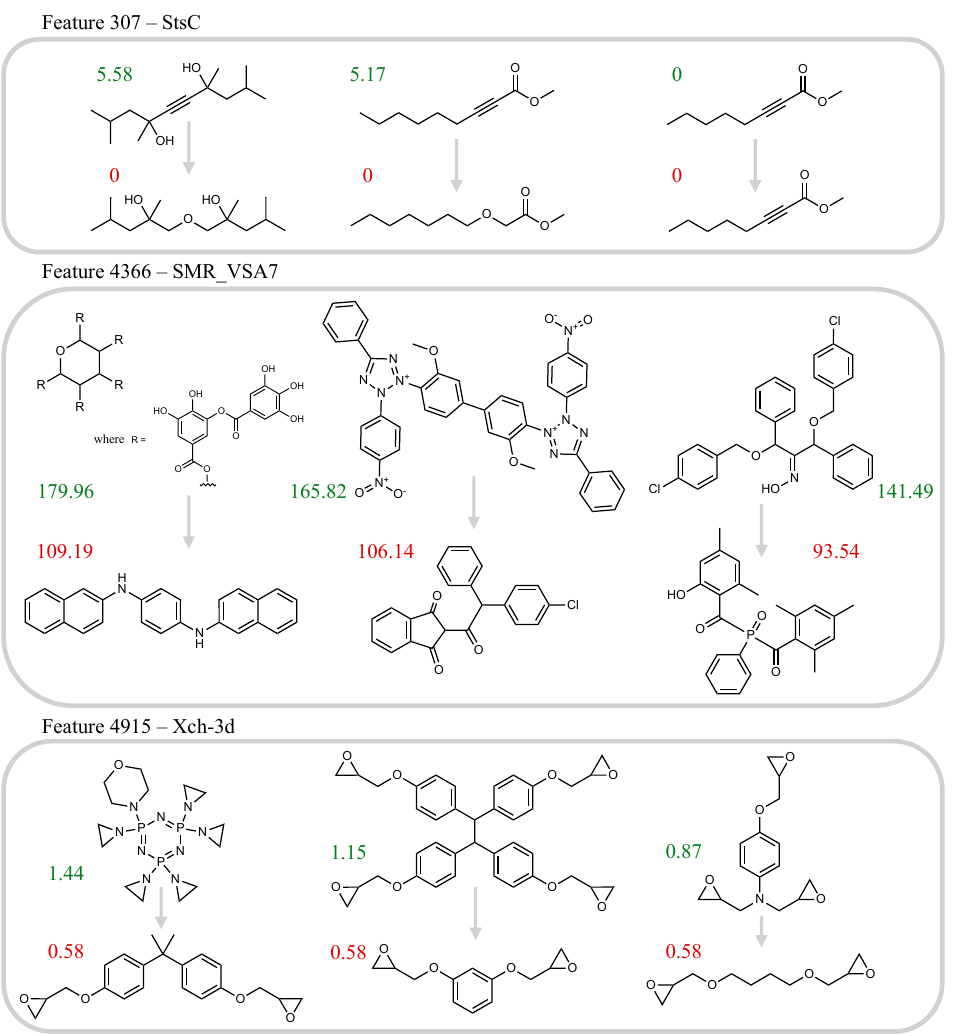}
    \caption{Top three activated molecules for the top three feature-descriptor relationships: StsC, SMR\_VSA7, Xch-3d. Molecules are then steered by setting the specified feature activation to 0, and the descriptor value is provided before (green) and after (red).}
    \label{fig:mols_physico}
\end{figure}

\FloatBarrier
\newpage
\subsection{Functional Behaviour}
\label{app:func-behaviour}

ChEMBL35 was subset to molecules with at least one Ki - ChEMBL target pair. For each feature, the molecules for which the feature had a normalised activation $> 0.5$ were selected. The set intersection of ChEMBL targets for this selection was calculated. Then the largest set was retrieved.

To establish a null distribution for assessing Tanimoto similarity (using radius-2, 4096-bit FeatureAtomInv Morgan fingerprints, with chirality), we first estimate the required sample size of random molecular pairs.

Given the equation for estimating a proportion (or similarity) with a given margin of error $\epsilon$ at a confidence level of $z$:
\begin{align}
    m &= \frac{z^2\sigma^2}{\epsilon^2}
\end{align}

\begin{figure}[h!]
    \centering
    \includegraphics[width=\linewidth]{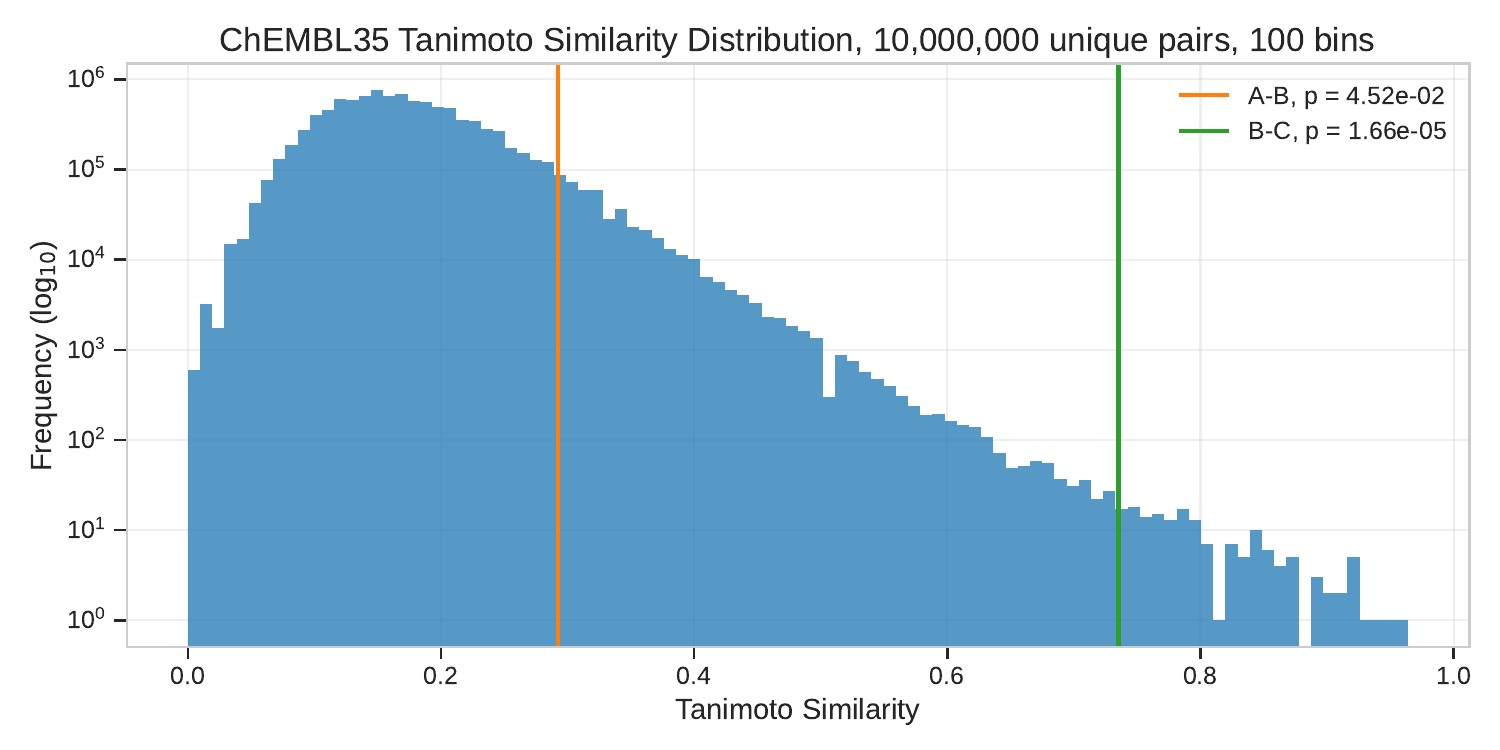}
    \caption{Distribution of Tanimoto Similarity (TS) across 10,000,000 unique pairs, sampled from ChEMBL35. The TS of CHEMBL1672485 (A) and CHEMBL454618 (B) (A-B: 0.293) is shown in orange, the TS of B and CHEMBL368522 (C) (B-C: 0.735) is shown in green. The corresponding $p$-values drawn from a right-tailed empirical distribution.}
    \label{fig:tanimoto_dist}
\end{figure}

Using a pilot study variance estimate of $\hat\sigma = 0.062$, from 100,000 pairs a desired margin of error of $\epsilon=0.0001$, and a 99\% confidence level, we calculate a required sample size of approximately 2.6 million pairs. We subsequently sample 10 million random molecular pairs from the ChEMBL dataset to construct a high-resolution empirical null distribution of Tanimoto similarity scores. This distribution serves as the baseline for computing the statistical significance of observed similarity values (see \Cref{fig:tanimoto_dist}).

\begin{figure}[h!]
    \centering
    \includegraphics[width=\linewidth]{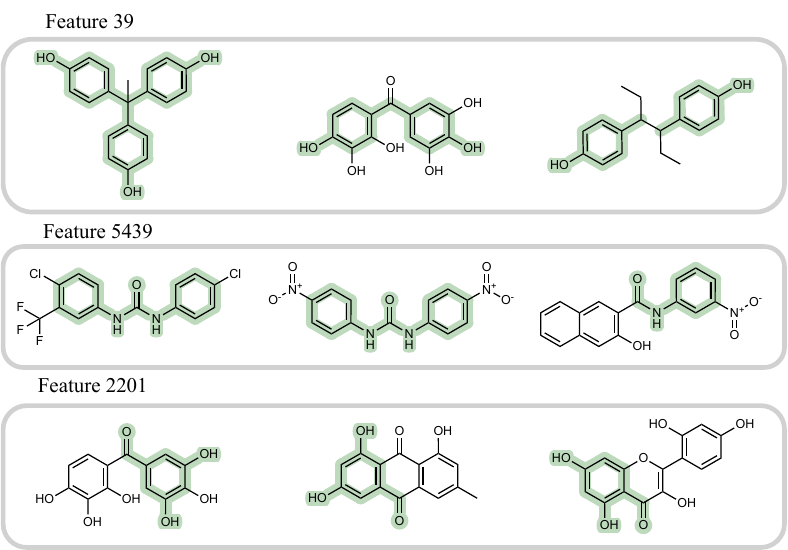}
    \caption{Top three activated molecules for the top three features used for logisitic regression of toxicity (see Section \ref{sec:results-func}). The major common substructure for each feature is highlighted in green.}
    \label{fig:mols_toxic}
\end{figure}

\begin{figure}[h!]
    \centering
    \includegraphics[width=\linewidth]{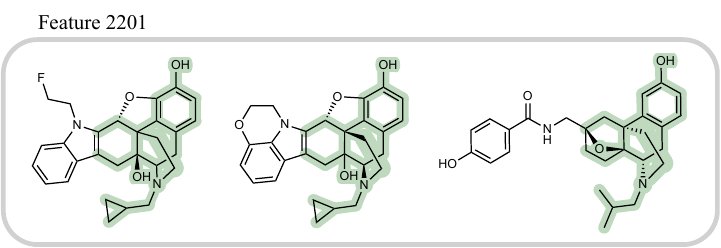}
    \caption{Top three activated molecules for the feature found to be related to pharmacological function to opioid receptors (see Section \ref{sec:results-func}). The major common substructure is highlighted in green.}
    \label{fig:mols_opioid}
\end{figure}

\newpage
\FloatBarrier
\section{Steering Stability}

To evaluate the stability and causal influence of the learned representations, we conduct a series of ablation experiments comparing the original SMI-TED neuron activations (dense representations) with our SAE-derived features (sparse representations). The experiments are performed on a 10,000-molecule subset of the MOSES dataset.

For the dense neuron representations, we intervene on each of the 768 neurons individually. For each neuron, we identify the 100 molecules that elicit its highest and lowest activations. Assuming a normal distribution of activations for a given neuron, we ablate its value to the distribution's mean for these selected molecules before decoding them back to SMILES strings. This intervention results in minimal change: only 14 of the 768 neurons produce any invalid SMILES upon ablation, and no interventions result in a valid but different molecule. This suggests the dense representations are highly robust, with chemical information distributed across many neurons, making individual neurons non-critical for reconstruction.

\begin{figure}[h!]
    \centering
    \includegraphics[width=\linewidth]{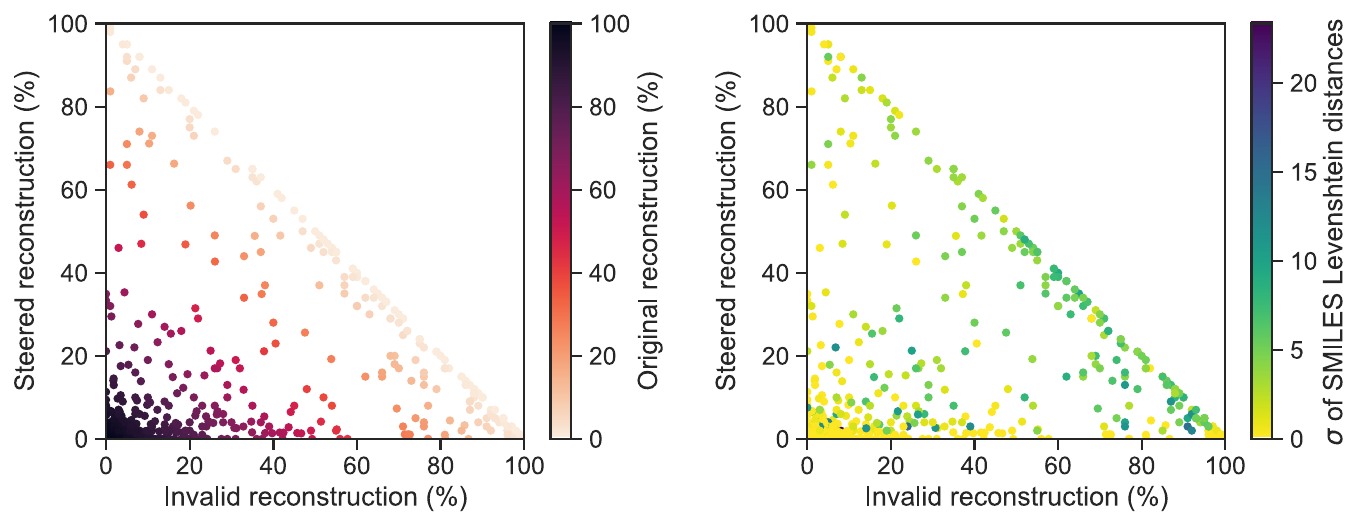}
    \caption{Reconstruction rates of 100 molecules after ablation of 2501 active features on a 10k subset of MOSES. After ablation of a feature, embeddings are decoded into either the original SMILES ("Original", hue (left)), a different SMILES ("Steered", y-axis), or an invalid SMILES ("Invalid", x-axis). The variability in steered transformations are shown by the standard deviations of Levenshtein distances between the original and steered SMILES strings (hue, right).}
    \label{fig:feature_steering}
\end{figure}

For the sparse SAE features, we perform a different intervention tailored to their sparse nature. For each of the 2,501 active features, we select up to 100 molecules where that feature has a non-zero activation. We then ablate the feature by setting its activation to zero, effectively ``turning it off,'' before decoding. The outcomes are then categorised as: 1) \textit{Original}: the decoded SMILES matches the original, 2) \textit{Invalid}: the decoded SMILES is chemically invalid, or 3) \textit{Steered}: the decoded SMILES is valid but different from the original.

In contrast to the dense neurons, the sparse features demonstrate significant steerability. We find that interventions on 749 of the 2,501 active features successfully steer molecules to new, valid chemical structures. As shown in \Cref{fig:feature_steering}, this approach reveals a clear trade-off between feature stability (valid reconstruction) and steerability, confirming that individual SAE features often represent specific, manipulable chemical concepts. Levenshtein distances between the original and steered SMILES strings are also used as an approximation for measuring the consistency of steering transformations for a given feature~\citep{rootaligned}. The Levenshtein distance is a metric for measuring the difference between two string sequences. As expected, the most stable and steerable features have a low standard deviation of Levenshtein distances, indicating that steering most likely changes most molecules in the same way. Features with a higher invalid reconstruction rate also have a higher standard deviation, indicating that their steered changes are less consistent.

\end{document}